\documentclass[final]{nesy2026} 

\usepackage{amsmath}
\usepackage{amssymb}
\usepackage{mathrsfs,amsopn}
\usepackage{stmaryrd}
\usepackage{multirow}
\usepackage{booktabs}
\usepackage{xcolor}
\usepackage{colortbl}

\usepackage[hide]{ed}

\usepackage{quiver}
\usepackage{tikz}
\usetikzlibrary{er,positioning}
\usepackage[most]{tcolorbox}
\newtcolorbox{impl}[2]{%
  blanker, breakable,
  left=0pt, right=0pt, top=0pt, bottom=0pt,
  overlay unbroken and first={%
    \node[anchor=north east, font=\footnotesize\ttfamily, inner sep=1pt, fill=white]
      at (frame.north east) {#2};}
}
\usepackage{minted}
\setminted{fontsize=\scriptsize, breaklines=true, breakanywhere=true, tabsize=2}

\usepackage{macros}

\newcommand{\nesycattorch}{\textsc{NeSyCat Torch}\xspace}

\usepackage{longtable}

\iffinalsubmission
  \jmlrproceedings{Preprint}{Preprint}%
  \jmlrvolume{}%
  \jmlrworkshop{}%
  \editors{}%
\fi

\title[\nesycattorch]{\nesycattorch: A Differentiable Tensor Implementation of Categorical Semantics for Neurosymbolic Learning}

\clearauthor{\Name{Daniel {Romero Schellhorn}} \Email{daniel.schellhorn@uni-osnabrueck.de}\\
  \Name{Till Mossakowski} \Email{till.mossakowski@uni-osnabrueck.de}\\
  \Name{Bj{\"o}rn Gehrke} \Email{bjoern.gehrke@uni-osnabrueck.de}\\
  \addr University of Osnabr{\"u}ck, Osnabr{\"u}ck, Germany}

\begin{document}

\maketitle

\begin{abstract}
Neurosymbolic semantics is fragmented: classical, fuzzy, probabilistic and neural systems each define truth by their own inductive rules. NeSyCat, extending ULLER, subsumes them under a single inductive definition of truth, parametric in a strong monad and an aggregation structure on truth-values. NeSyCat has so far lacked an account of predicates and functions learned by neural networks. We provide \nesycattorch as the missing link and interpret computational symbols via neural networks, implementing the framework in probabilistic programming and tensor-based backends. We use the distribution monad for reference semantics and metric evaluation, and complement it by a monad for numerically stable, differentiable training: the lazy log-tensor monad over the log-semiring. For efficient training in batches, we furthermore employ a batch monad. The axioms \emph{are} the source code: written once in monad-based \doNotation, monadic bind performs marginalisation, lazily pruning unneeded branches. On MNIST addition, our HaskTorch, JAX, and PyTorch implementations outperform LTN and DeepProbLog in speed and accuracy, while achieving nearly the accuracy of DeepStochLog. However, unlike DeepStochLog, we stay in a uniform framework that applies to many first-order NeSy approaches. Namely,
the construction is parametric in the monad; instantiating it with, e.g., the Giry monad extends the approach to continuous probability (working out a neural representation here is left for future work).
\end{abstract}

\section{Introduction}
\label{sec:intro}

Neurosymbolic (NeSy) AI combines the perceptual strength of neural networks with the structured,
verifiable reasoning of symbolic logic. A recurring obstacle is fragmentation: classical, fuzzy,
and probabilistic NeSy systems each come with their own logical language and semantics, so knowledge
bases and learning objectives rarely transfer between them. ULLER - the Unified Language for Learning
and Reasoning \citep{vankriekenULLER2024} -  endows First-Order-Logic (FOL) syntax with three pairwise-independent
semantics - classical, fuzzy, probabilistic - each carrying its own inductive definition of truth.

A recent line of work \citep{schellhornNeSyCatCategorical2026} reformulates all three
semantics as instances of a single \emph{categorical} framework built on \emph{monads}, Moggi's
construct for computational effects in functional programming
\citep{moggiNotionsComputation1991}. The key observation is that an ULLER computation formula
$x := m(T_1,\dots,T_n)\,(F)$, interpreted as ``run model $m$, then bind its result to $x$, then evaluate $F$'', is exactly monadic \textbf{do}-notation. Fixing a strong monad $\mathcal M$ (the
effect) and an aggregated truth-value space $\Omega$ with connectives and quantifiers yields a
\emph{NeSy framework}; classical, fuzzy, probabilistic, LTN, and
possibilistic semantics all reappear as choices of $\mathcal M$ and $\Omega$, evaluated by
\emph{one} inductive definition of truth.

For efficiency reasons, we use \emph{lazy} monads. 
The lifting operation in the distribution monad computes probabilities using marginalization; a lazy monad ensures that marginalization is only done in cases where it is actually needed.

Besides usual FOL function and predicate symbols, we consider computational function symbols $X\to \mathcal M Y$ and computational predicate symbols $X\to \mathcal M \Omega$. At the deep learning level, we need also to consider two-sided computational function symbols $\mathcal M X\to \mathcal M Y$ and two-sided computational predicate symbols $\mathcal M X\to \mathcal M \Omega$. 

We now recall monads and present the monads used in this paper in Table~\ref{tab:monads-nesycat}:

\section{Monads for Computational Effects}
\label{sec:monads}

\begin{definition}[Monad {\citep[\S 3.1]{kohlSchwaigerMonads2021}}]
A monad is given by a triple \minth{(m, return, >>=)} where \minth{m} is a type constructor mapping a type \minth{a} to a type \minth{m a} of computational effects with values from \minth{a}, \minth{return} embeds values into computation and
\minth{>>=} (bind) is used for composition of computations:
\begin{minted}{haskell}
return :: a -> m a
(>>=)  :: m a -> (a -> m b) -> m b
\end{minted}
Here, $c$ \minth{>>=} $f$ first executes computation $c$ over type \minth{a} and passes its value(s) to a function $f$ delivering a computation over type \minth{b}.
A monad needs to satisfy associativity and unit laws. 
\end{definition}

\noindent
Haskell provides the \minth{do} notation as syntactic sugar for composing monadic maps: with it monadic code almost looks like imperative
code, but under the hood there are only pure maps and monads: \minth{do { x <- y; f}} is syntactic sugar for \minth{y >>= (\x -> do { f })}.

\begin{table}[t]
  \centering\small
  \begin{tabular}{@{}cclp{\dimexpr\textwidth-6.6cm\relax}@{}}
  \toprule
  \textbf{Monad} & \textbf{Code} & \textbf{Effect} & \textbf{Description} \\
  \midrule
  $\mathcal D$ & \minth{Dist} & finite probability &
    Finitely supported probability distributions; the reference semantics and metric readout. \\[3pt]
  $\mathcal T$ & \minth{Tens} & logit weights &
    Finite-support Tensor monad $\mathcal{T}\,m = \mathbb R^m$ (leaves are weight tensors) and
    \minth{(>>=)} is the linear pushforward. \\[3pt]
  $\mathcal{T}_{\log}$ & \minth{LogTens} & stable arithmetic &
    \minth{Tens} in logarithmic coordinates, over the log-semiring
    $(\mathbb R, \mathrm{logsumexp}, +)$: numerically stable and differentiable, the monad used in training.
    \\[3pt]
  $\mathcal B$ & \minth{Batch} & batching &
    Reader monad on the batch index object $\underline B$ (Sec.~\ref{sec:training}), parallel
    processing of a mini-batch in training. \\
  \bottomrule
  \end{tabular}
  \caption{Monads in \nesycattorch, in the format of \citet[Table~1]{kohlSchwaigerMonads2021}.}
  \label{tab:monads-nesycat}
  \end{table}

\paragraph{Three monad layers.} In this work, we will use \emph{three} different monads:
\begin{enumerate}
\item the \emph{probability layer}: the distribution monad $\mathcal D$ \citep{schellhornNeSyCatCategorical2026} is the reference semantics. ${\mathcal M} X$ is the space of all finitely supported probability distributions over $X$. \minth{return} $x$ is the distribution assigning all probability mass to $x$. $f$ \minth{>>=} $\rho = \sum_{x\in X}\!f(x)(y)\cdot\rho(x)$. This corresponds to a two-level random process: first $x$ is drawn from $\rho$, then $y$ is drawn from $f(x)$. This results in a marginal distribution for the joint distribution $\hat\rho(x,y):= f(x)(y) \cdot \rho(x)$.
 Distributions over Booleans can be regarded as single
  probabilities (namely those for $\mathit{true}$). 
\item the \emph{tensor layer}: the tensor monad $\mathcal{T}$ works with the same equations. It implements the distribution monad in a differentiable way, but uses real numbers (logits) instead
  of probabilities. It is connected to $\mathcal D$ by a bridge (see Sect.~\ref{sec:mnist} below). The implementation uses \minth{LogTens}, i.e.\ $\mathcal{T}$
  in \emph{logarithmic coordinates} (over the log-semiring) for stable differentiation.
\item the \emph{batching layer}: the batch monad $\mathcal B$, defined in Sect.~\ref{sec:training} below, allows the parallel
  processing of training samples, evaluated together in one run. It carries no probability and no
  geometry, and, dually to the first layer, it never uses the samples' randomness.
\end{enumerate}

\section{Syntax and Semantics}
\label{sec:syntax-semantics}
Using these monads, we extend the categorical semantics of \citet[D1.1.]{johnstoneElephant2002} by adding monad symbols and monadic interpretations.
Refining \citep{schellhornNeSyCatCategorical2026}, \nesycattorch organises syntax and semantics into four layers,
each a pair of a \emph{signature} (only symbols) and an \emph{interpretation} (their meaning):
categorical, logical, domain, and grammatical. The first three declare and interpret
symbols; the fourth generates terms and formulas over them and assigns them their monadic
semantics by one induction.

\subsection{Categorical, logical and domain layers}
\label{sec:layer-categorical-logical}

(i) Categorically, we work in the category of sets and maps, $\Set$; computationally it is the category of
(inhabited) Haskell types and maps, on which every Haskell monad of
Section~\ref{sec:monads} is defined. Parameters are drawn from the category $\Tens$ of tensor spaces, which has as objects the Euclidean spaces $\mathbb R^n$ and as morphisms differentiable maps. Implementation-wise, these are implemented as HaskTorch/PyTorch/JAX tensors, where the neural networks live and backpropagation is performed \citep{fongBackpropFunctor2019}. 

The one choice that genuinely varies is the
\emph{effect}: we keep a single \textbf{monad symbol} $\bigcirc$, which allows for the choice of a monad $\mathcal M := \mathcal I(\bigcirc)$ from Table~\ref{tab:monads-nesycat}. This paper uses two readings: \emph{probabilistic}
($\mathcal M := \mathsf{Dist}$), and \emph{tensorial} ($\mathcal M := \mathcal{T}$).

(ii) Logically, the basic truth values are Booleans, $\Omega := \mathbb B = \{\mathbf{False},\mathbf{True}\}$. The
\textbf{connective symbols} $\Conn$ (e.g.\ $\wedge$, $\vee$, $\neg$, $\rightarrow$) are the ordinary
Boolean operations \citep[D1.1]{johnstoneElephant2002} given by $\mathcal I({*}) \colon \mathbb B^{\ari(*)} \to \mathbb B$.
In the neural setting, though, a predicate does not return a plain Boolean but a \emph{monadic} truth value
in $\mathcal M\,\Omega$, for example a distribution over $\mathbb B$, or a vector of logit weights. Hence we lift each connective
to act on $\mathcal M\,\Omega$: bind its arguments, apply the Boolean operation to the plain truth-values
and then return the result (connective clause of Def.~\ref{def:Semantics}).

The only
logical symbols whose interpretation is genuinely monadic are the \textbf{quantifier
symbols} $\mathrm{Quan}$: a quantifier $Q$ takes its body (a map into monadic truth) and aggregates it to a monadic truth value, indexed by a finite (infinitary are future work) object $D$:
\[
  \mathcal I(Q)_D \colon (D \to \mathcal M\,\Omega) \;\longrightarrow\; \mathcal M\,\Omega,
\]

\begin{definition}[Domain signature $\Sigma$]
A domain signature consists of \textbf{domain symbols} $\mathrm{Dom}$, \textbf{variable symbols} $\Var$ (each over a domain symbol,
$x : S$), and \textbf{function symbols} $\Fun$ and \textbf{relation symbols} $\Rel$, each
partitioned into
$\Fun = \Fun^{\mathrm{Tarski}} \sqcup \Fun^{\mathrm{Kleisli}} \sqcup \Fun^{\mathrm{Nesy}}$ and $\Rel = \Rel^{\mathrm{Tarski}} \sqcup \Rel^{\mathrm{Kleisli}} \sqcup \Rel^{\mathrm{Nesy}}:$
\begin{center}\small
\begin{tabular}{@{}llll@{}}
\toprule
\textbf{Kind} & \textbf{Function symbol} & \textbf{Relation symbol} & \textbf{Implementation} \\
\midrule
Tarski  & $f : S_1,\ldots,S_n \to T$ & $R : S_1,\ldots,S_n \to \tau$ & deterministic \\[2pt]
Kleisli & $f : S_1,\ldots,S_n \to \bigcirc T$ & $R : S_1,\ldots,S_n \to \bigcirc\tau$ & effectful \\[2pt]
Nesy    & $f : \bigcirc S_1,\ldots,\bigcirc S_n \to \bigcirc T$
        & $R : \bigcirc S_1,\ldots,\bigcirc S_n \to \bigcirc\tau$ & neural \\
\bottomrule
\end{tabular}
\end{center}
Here, $\tau$ is the type of truth values.
In order to stay close to FOL, we disallow function and relation symbols taking arguments of type $\tau$ or $\bigcirc\tau$.
\end{definition}

\begin{definition}[Domain interpretation $\mathcal I$] A domain interpretation assigns a set
$\mathcal I(S)$ to each domain symbol $S$, and to each function symbol $f$ a map $\mathcal I(f)$ as in
\begin{center}\small
\begin{tabular}{@{}ll@{}}
\toprule
\textbf{Signature} & \textbf{Interpretation} \\
\midrule
$f : S_1,\ldots,S_n \to T$
  & $\mathcal I(f) \colon \mathcal I(S_1,\ldots,S_n) \to \mathcal I(T)$ \\[3pt]
$f : S_1,\ldots,S_n \to \bigcirc T$
  & $\mathcal I(f) \colon \mathcal I(S_1,\ldots,S_n) \to \mathcal M\,\mathcal I(T)$ \\[3pt]
$f : \bigcirc S_1,\ldots,\bigcirc S_n \to \bigcirc T$
  & $\mathcal I(f) \colon \mathcal M\,\mathcal I(S_1) \times \cdots
     \times \mathcal M\,\mathcal I(S_n) \to \mathcal M\,\mathcal I(T)$ \\
\bottomrule
\end{tabular}
\end{center}
where $\mathcal M := \mathcal I(\bigcirc)$ and
$\mathcal I(S_1,\ldots,S_n) := \mathcal I(S_1) \times \cdots \times \mathcal I(S_n)$. Relations similarly, with $\mathcal I(T)$ replaced by $\Omega$. A Tarski symbol is a plain map, a Kleisli symbol a map into
$\mathcal M\,\mathcal I(T)$, and a Nesy symbol takes monadic carriers as input and produces monadic carriers as output.
\end{definition}

\subsection{Grammar}
\label{sec:layer-grammatical}

We write terms $t$ in the \doNotation  of
Section~\ref{sec:monads}, the same notation the implementation uses. To avoid duplication of \doNotation for terms and formulas, we construe formulas $\phi$ as terms of type $\tau$ and connectives as operations on $\tau$.\footnote{This is standard in higher-order logic \citep{church1940formulation,Andrews86}, and with suitable restrictions, it can be used also for first-order logic.} Tarski terms are terms of type $S$ or formulas of type $\tau$. Monadic terms are terms of type $\bigcirc S$ or $\bigcirc \tau$ (the latter are monadic formulas); such terms can be used as terms $t_1,\ldots,t_n,t$ in the \DO-notation. In order to stay close to FOL, the term $t$ in $\mathbf{return}\ t$ must have a Tarskian, i.e.\ $\bigcirc$-free type.
With
$x \in \Var$, $c,f \in \Fun\cup\Rel\cup\Conn$, $Q \in \mathrm{Quan}$, we define terms as follows:
\[
  \begin{aligned}
    t \;&::=\; x \;\mid\; c \;\mid\; f(t_1,\ldots,t_n) && \text{(standard FOL term/formula)}\\
    &\;\mid\; \mathbf{return}\ t \;\mid\;  \DO\{\, x_1 \leftarrow t_1;\, \ldots;\, x_n \leftarrow t_n;\ t \,\} 
    && \text{(monadic term/formula)}\\
    &\;\mid\; Qx(t) && \text{(quantifier, $t,Qx(t):\bigcirc \tau$)}\\
  \end{aligned}
  \]
  Each term $t$ has free variables $\inn(t)$ and (result) type $\out(t)$.
  Term formation has to be type-correct.
  Formulas $\phi$ are terms with $\out\phi=\tau$.
Standard FOL terms can have type $S$ (or $\tau)$ or $\bigcirc S$ (or $\bigcirc \tau$), depending on whether $f$ has a monadic result. Monadic terms are always of type $\bigcirc S$ (or $\bigcirc \tau$).
\doNotation has to be used for the application of connectives to monadic formulas.  
We can recover the original ULLER and NeSyCat syntax for terms $T_i$, formula $F$ with variable $x$, and a neural model $m$ in \nesycattorch as follows:
$$[x:=m(T_1,\dots,T_n)]F \quad \equiv \quad \DO \{\ x\leftarrow m(T_1,\dots,T_n);\; F\ \} $$
\noindent
We define the semantics \emph{pointwise} at a 
valuation $\nu$: for the context $[x_1{:}S_1,\ldots,x_k{:}S_k]$, $\nu$ maps variables to values and can be construed as a tuple in
$\mathcal I(S_1) \times \cdots \times \mathcal I(S_k)$, one component per variable. 

The following semantics uses the \doNotation of Section~\ref{sec:monads}.
Subterm values are \emph{always} monadic and therefore always bound with
$\leftarrow$. A \emph{Tarski} symbol is pure, so its result re-enters the monad through
\minth{return}; a \emph{Kleisli} symbol is already monadic, so it stands as the final
\doExpression without \minth{return}; a \emph{Nesy} symbol takes the monadic values
themselves, so no binds occur at all. A quantifier likewise consumes its body as the map $a \mapsto \sem{\phi}(c,a)$ directly, matching its interpretation type from
Section~\ref{sec:layer-categorical-logical}.

\begin{definition}[Semantics $\sem{\cdot}$]
\label{def:Semantics}
Terms and formulas denote maps
\[
  \sem{t} \colon \mathcal I(\inn(t)) \to \mathcal I(\out(t)),
\]
defined by induction on the grammar with $\vec t:=(t_1, \ldots, t_n)$: 

{\setlength{\abovedisplayskip}{4pt}\setlength{\belowdisplayskip}{4pt}%
 \setlength{\abovedisplayshortskip}{4pt}\setlength{\belowdisplayshortskip}{4pt}
\begin{align*}
  \sem{x}(\nu) \;&=\; \nu(x)
    &\qquad
  \sem{f(\vec t)}(\nu) \;&=\;
    \mathcal I(f)\bigl(\sem{t_1}(\nu),\, \ldots,\, \sem{t_n}(\nu)\bigr)
    \\[2pt]
  \sem{c}(\nu) \;&=\; \mathcal I(c)
    &\qquad
  \sem{\mathbf{return}\ t}(\nu) \;&=\;
    \mathbf{return}\ \sem{t}(\nu)
\end{align*}
\begin{align*}
  \sem{\DO\{\, x_1 \leftarrow t_1;\, \ldots;\, x_n \leftarrow t_n;\ t \,\}}(\nu) \;&=\;
    \DO\{\, a_1 \leftarrow \sem{t_1}(\nu);\,\ldots;\, a_n \leftarrow \sem{t_n}(\nu);\\
  &\hphantom{{}=\;}\qquad \sem{t}(\nu, x_1\mapsto a_1,\ldots x_n\mapsto a_n)
     \,\}
\end{align*}
\begin{align*}
  \sem{Qx(t)}(\nu) \;&=\;
    \mathcal I(Q)_{\mathcal I(S)}\bigl(\lambda a.\,\sem{t}(\nu, x\mapsto a)\bigr) \qquad (x : S)
\end{align*}}
\end{definition} 

\noindent
The order of independent binds in these clauses is irrelevant: all monads in this paper are
\emph{commutative} \citep{kockMonadsSymmetric1970} ($\mathsf{Dist}$: the independent joint
distribution; $\mathcal{T}$: the outer product), so commutative
connectives stay commutative under the monadic reading.
Section~\ref{sec:mnist} unfolds these clauses step by step on the running example:

\section{The Running Example: MNIST Addition}
\label{sec:mnist}

We instantiate the MNIST single-digit addition under \emph{distant supervision}: only the
\emph{sum} of two handwritten digits is observed, never the digits. The axiom is
$\forall (x,y,n){:}S \, \bigl(n \mathrel{\mathsf{=}} \mathsf{digit}(x) \mathrel{\mathsf{+}} \mathsf{digit}(y)\bigr)$.
The domain signature $\Sigma$ has sorts $\mathsf{Image},\mathsf{Digit},\mathsf{Nat}$, one
Kleisli symbol $\mathsf{digit}:\mathsf{Image}\to\mathcal M\,\mathsf{Digit}$, and two Tarski
symbols $\mathsf{+}:\mathsf{Digit}^2\to\mathsf{Nat}$, $\mathsf{=}:\mathsf{Nat}^2\to\tau$. Writing
$\mathrm{dig}_\theta:=\mathcal I_\theta(\mathsf{digit})$, the clauses of
Section~\ref{sec:layer-grammatical} are given as:
\[
  \begin{aligned}
\mathbf{do}\ \bigl\{\, d_1 \leftarrow \mathrm{dig}_\theta(x);\ d_2 \leftarrow \mathrm{dig}_\theta(y);\
       \mathbf{return}\,\bigl(n \mathrel{{=}} (d_1 \mathrel{{+}} d_2)\bigr) \bigr\}.
  \end{aligned}
\]
\noindent
Each $\leftarrow$ performs the law of total
probability in $\mathcal{D}$ and the log-space convolution in $\mathcal{T}$ respectively, so this
program is not pseudo-code: it \emph{is} the semantic value, the monad laws guarantee it
equals the literal inductive unfolding, and the identical text compiles as Haskell.

\paragraph{What ``$+$'' becomes.} The Tarski symbol $\mathsf{+}$ is ordinary addition
$\mathsf{sum}:\underline n\times\underline m\to\underline{n{+}m{-}1}$ in the base, where $\underline k=\{0,\dots,k-1\}$; the monad enters
only by \emph{lifting} it: the functor sends this map to $\mathcal{T}(\mathsf{sum})\colon\mathcal{T}(\underline n\times\underline m)\to\mathcal{T}(\underline{n{+}m{-}1})$. Over $\mathbb R$ the carrier is the finitely supported tensor monad
$\mathcal{T}\,m=\mathbb R^m$ (unit $\eta_m(i)=e_i$), a lawful Haskell
\minth{Monad}.\footnote{The finite-dimensional vector-space construction is, strictly, a
\emph{relative} monad on $\mathbf{Fin}\hookrightarrow\mathbf{Set}$ rather than an
endofunctor \citep{altenkirchMonadsNotEndofunctors2015}: its bind sums only over finite index set. The finitely supported version is an ordinary monad.} This lift is the pushforward, which on $\mathsf{sum}$
marginalises a joint vector onto the sum. The two classifiers return a \emph{pair}, joined by the outer product $a\otimes b$, yielding
exactly the discrete convolution:
\[
  \mathcal{T}(\mathsf{sum})(a\otimes b)(s) = \textstyle\sum_{i+k=s} a_i\,b_k = (a * b)(s),\qquad s\in\{0,\dots,18\},
\]
i.e.\ the unnormalized distribution of the sum of two independent digits, exactly what the axiom scores against
(the DeepProbLog reading \citep{DBLP:journals/ai/ManhaeveDKDR21}).

\paragraph{The first two monad layers.}  The digit symbol is \emph{two-sided},
$\mathrm{dig}_\theta \colon \bigcirc\mathsf{Image} \to \bigcirc\mathsf{Digit}$: an observed image enters
as a certain one-hot encoded tensor. Why encode a tensor that is already a tensor? Because the input lives in
$\bigcirc\mathsf{Image}$, a \emph{distribution} over images, and a single observation is just the certain
case: the point mass at that image (one-hot in $\mathcal{T}$, a delta distribution in $\mathcal D$), exactly how one
datum is represented in an empirical data distribution as empirical state. The encoding sends a delta distribution on an element to the one-hot encoding on the same element. The decoding sends logit tensors via a softmax to a distribution. The
probability reading is \emph{defined} as the composition
$\mathrm{dig}_\theta^{\mathcal D} = \mathrm{dec}\circ\mathrm{dig}_\theta^{\mathcal{T}}\circ\mathrm{enc}$:
\[\begin{tikzcd}[ampersand replacement=\&, sep=large]
  {\mathcal{T}(\mathsf{Image})} \&\& {\mathcal{T}(\mathsf{Digit})} \\
  {\mathcal D(\mathsf{Image})} \&\& {\mathcal D(\mathsf{Digit})}
  \arrow["{\mathrm{dig}_\theta^{\mathcal{T}}\ (\text{CNN})}", from=1-1, to=1-3]
  \arrow["{\mathrm{dig}_\theta^{\mathcal D}}"', from=2-1, to=2-3]
  \arrow["{\mathrm{enc}}", from=2-1, to=1-1]
  \arrow["{\mathrm{dec}=\mathrm{softmax}}", from=1-3, to=2-3]
\end{tikzcd}\]
  
 \noindent 
The \textbf{do}-notation derivation above is no
paraphrase: it is the actual Haskell, written once and polymorphic over the monad $m$. The sorts are
plain types, $\mathtt{+}$ and $\mathtt{==}$ are host functions, and
\minth{digit} is the only monad-dependent symbol and the bind
supplies the marginalisation. In Python, we use \mintp{yield} instead of \minth{<-} utilising the generator syntax as syntactic sugar for the same monadic composition as Haskell's \textbf{do}-notation.

\begin{impl}{The abstract formula, written once (monad-polymorphic)}{}
  \small
  \begin{minted}{python3}
class MNistAddition(Example, DistLogTensBridge):
    def formula(self, m, x: Monad[Image], y: Monad[Image], n: Monad[int]) -> Formula[bool]:
        d1 = yield self.digit(m, x)
        d2 = yield self.digit(m, y)
        s = yield n
        return s == d1 + d2
  \end{minted}
\end{impl}

\noindent
Here, apart from the bind \mintp{s = yield n}, which relates only to batching and is explained in Section~\ref{sec:training}, the formula corresponds to the abstract formula in Section~\ref{sec:layer-grammatical}.
The arithmetic is interpreted identically in both monads - plain integer addition - and the only
per-monad choice is how \minth{digit} is read: in \minth{LogTens} the CNN's raw logits become a
leaf; in \mintp{Dist} that same leaf is decoded to a distribution:

\begin{impl}{Two interpretations of \mintp{digit} (one per monad)}{}
  \small
  \begin{minted}{python3}
class MNistAddition(Example, DistLogTensBridge):
    @monad_method
    def digit(self, img: Monad[Image]) -> Monad[int]: ...

    @digit.instance(LogTens)
    def digit_logtens(self, img: LogTens[Image]) -> LogTens[int]:
        model = self.tensor_interpretation.models[type(self).digit]
        return LogTens.bind(img, lambda x: LogDefer(list(range(10)), x, model))

    @digit.instance(Dist)
    def digit_dist(self, img: Dist[Image]) -> Dist[int]:
        return self.decode(self.digit(LogTens, self.enc_dist(img)))
  \end{minted}
\end{impl}

\noindent
The \minth{LogTens} monad represents the type constructor \minth{Tens a = [a -> Real]}
realized over the log-semiring $(\mathbb R,\mathrm{logsumexp},+)$ with finite support given by a list \minth{[a]} of elements of type \minth{a}:

\begin{impl}{The differentiable log-tensor monad (a free monad)}{}
  \small
  \begin{minted}{python3}
class LogTens[A](Monad[A]):                  class Pure[A](LogTens[A]):
    ...                                          value: A

class Bind[A, B](LogTens[A]):                class LogLeaf[A](LogTens[A]):
    dist: LogTens[B]                             support: list[A]
    func: Callable[[B], LogTens[A]]              log_weights: torch.Tensor
  \end{minted}
\end{impl}

\noindent
\minth{LogTens} is used for training since log space is computationally convenient: products of probabilities become sums and marginalisation becomes
log-sum-exp, the numerically stable evaluation \citep{blanchard2021accurately}; weights
range over all of $\mathbb R$ instead of being squashed into $[0,1]$, where
differentiable fuzzy operators degenerate gradient-wise
\citep{kriekenDifferentiableFuzzyOps2022}; and, since softmax is shift-invariant
($\mathrm{softmax}(z) = \mathrm{softmax}(z + c\,\mathbf 1)$), normalising early would
discard the shift, so it is deferred - score now, normalise once at the boundary.

\section{Training with Monads}
\label{sec:training}

Training minimises the \emph{knowledge loss}, the mean negative log-truth of axioms over data,
\[
  \widehat L(\theta) \;=\; \tfrac{1}{N}\sum_{i=1}^{N} -\log \varphi_\theta(s_i),
\]
where $\varphi_\theta(s)$ is the truth value the axiom's semantics assigns to sample $s$
under weights $\theta$ (empirical risk minimisation; equivalently, maximum likelihood on the logical
evidence). This section answers the \emph{computational} question: why \emph{one} batched evaluation
of the axiom computes $\widehat L$, and in particular why an observation may be \emph{bound} in the
\doNotation. 

The obvious
way to compute $\widehat{L}$ runs the \doNotation program $N$ times, once per sample. The
implementation instead runs it on a whole \emph{batch} of samples at once, on tensors
carrying a batch axis.

The finite-sample average $\widehat L$ lives entirely in the first monad - it
exists because the data is a \emph{finite sample} of the world, not because of batching.

The per-sample formula is written once, polymorphic over the first two monad layers (in
$\mathcal D$ for evaluation, in $\mathcal{T}$ for training). The batching layer
is different: it is never a reading of the formula but is applied \emph{outermost}, the training monad is the composite
$\mathcal B \circ \mathcal{T}$.

\begin{definition}[Batch monad]
Let $\underline B = \{1,\dots,B\}$ index the samples evaluated together in \emph{one} run, a
mini-batch of the observations, with $B = N$ for the full sample. The \emph{batch monad}\footnote{The same repeated, conditionally independent structure is called a \emph{plate} in graphical models \citep{buntine1994operations}; cf.\ \texttt{pyro.plate} \citep{bingham2019pyro} and plated factor graphs \citep{obermeyer2019tensor}.} is the
$B$-fold power, i.e.\ the reader monad on $\underline B$:
\[
  \mathcal B\,X \;=\; X^{\underline B} \;\cong\; (\underline B \to X), \qquad
  \eta_{\mathcal B}(x) = \mathrm{const}\,x, \qquad
  (m \bind f)(b) \;=\; f\bigl(m(b)\bigr)(b).
\]
Its bind is the \emph{diagonal}: the continuation of index $b$ sees only index $b$'s value.
\end{definition}

In order to combine the batch monad with other monads, we need to extend it to a \emph{monad transformer}. Monad transformers transform a given monad to a new one by adding extra structure; in this case, by adding a batch dimension.
\begin{proposition}[Batch transformer]
For \emph{every} strong monad $\mathcal M$ the composite
\[
  \mathcal B \mathcal M\,X \;=\; \bigl(\underline B \to \mathcal M\,X\bigr)
\]
is again a monad (the reader monad transformer at $\underline B$ applied to $\mathcal M$), and there are two
canonical monad morphisms into it:
\[
  \mathrm{lift}_{\mathcal M} \colon \mathcal M\,X \to \mathcal B\mathcal M\,X,\;
  \mathrm{lift}_{\mathcal M} = \mathrm{const}
  \qquad\text{and}\qquad
  \mathrm{lift}_{\mathcal B} \colon \mathcal B\,X \to \mathcal B\mathcal M\,X,\;
  \mathrm{lift}_{\mathcal B}\ m = \eta_{\mathcal M} \circ m.
\]
$\mathrm{lift}_{\mathcal M}$ embeds a batch-constant effect;
$\mathrm{lift}_{\mathcal B}$ embeds a batched but \emph{certain} value.
\end{proposition}

A single run in the resulting composite provably yields all per-sample
values of its batch at once (Proposition~\ref{prop:pointwise}).
An observation is a value of the \emph{pure} batch monad, batched but certain,
$m \colon \underline B \to X$, and enters a formula only through
$\mathrm{lift}_{\mathcal B} = \eta_{\mathcal M} \circ (-)$. In the \doNotation this is the
bind $s \leftarrow m$: the bound $s$ is per-sample data carrying no uncertainty.

The training semantics of this paper is the composite $\mathcal B \circ \mathcal{T}$: a
leaf with log-weight tensor of shape $[B, k]$ is the carrier of $\underline B \to \mathcal{T}\,X$
under the \emph{rectangularity} invariant - all $B$ component measures of the batch share one support of
size $k$.
Under this representation $\mathrm{lift}_{\mathcal B}$ is the one-hot embedding: an observed
value becomes the $[B,k]$ one-hot log-weight leaf, the log-space image of $\eta_{\mathcal M}$
(the numerical floor at $\epsilon$ stands in for $\log 0 = -\infty$).

\section{Inference and Evaluation}
\label{sec:eval}

The inferential level turns the formula into an optimisation problem: the knowledge loss
$\widehat L$, a data loss $L_{\mathrm{data}}$, and their convex
combination. MNIST addition is
pure distant supervision: Adam minimises only the knowledge loss
$\widehat L(\theta)$.

We run the same specification in two backends, Haskell (HaskTorch) and Python (JAX). In Python JAX the log-space convolution is JIT-compiled and
differentiated. All numbers below come from the JAX backend on a single A100 GPU, averaged over $15$
seeds. We never use digit labels: the network learns to read the digits (about $97\%$ digit accuracy)
from the observed sums alone.

\begin{table}[h]
  \centering
  \caption{Single-digit MNIST addition: \nesycattorch's two variants, $3000$ training pairs. The DPL-style sweep is longer only because its harness uses batch $2$ (${\sim}10\times$ more steps) - the network is not slower, its per-step cost is even a little lower.}
  \label{tab:mnist}
  \small
  \begin{tabular}{@{}lcc@{}}
    \textbf{Metric} & \textbf{LTN-style} (100n. head, ELU) & \textbf{DPL-style} (120n. head, ReLU) \\
    \midrule
    Sum accuracy (test)                  & $94.6 \pm 0.6$\,\% & $94.2 \pm 0.7$\,\% \\
    Sum accuracy (train)                 & $\approx 100.0$\,\% & $\approx 100.0$\,\% \\
    Train / test step (batch $32$) & $0.52$\,/\,$0.20$\,ms & $0.44$\,/\,$0.20$\,ms \\
    Wall-clock, $15$-seed sweep  & $\approx 1.5$\,min & $\approx 5.6$\,min \\
    Training harness                     &$32$ batches, $20$ epochs & $2$ batches, to convergence. \\
  \end{tabular}
  \end{table}

Table~\ref{tab:addition} puts \nesycattorch next to LTN \citep{DBLP:journals/ai/BadreddineGSS22},
DeepProbLog \citep{DBLP:journals/ai/ManhaeveDKDR21}, and for two digits, logLTN
\citep{badreddineLogLTN2023}, NeuPSL and DeepStochLog. For the single-digit case we give each \nesycattorch
variant the same CNN, batch size and number of epochs as the baseline it is compared to. \nesycattorch is ahead in both matchups: $94.6$ vs LTN's $93.5$, and $94.2$
vs DeepProbLog's $92.2$. The real gap is a little larger, because LTN reports only its $10$ best of
$15$ runs (about one in five gets stuck early on), while we average all $15$. Although this issue has been solved by working in log space \citep{badreddineLogLTN2023}, we outperform even logLTN in both accuracy and speed. In general, speed is not even an issue: at batch $32$ our step times (Table~\ref{tab:mnist}) are well under the $5.36$/$3.44$\,ms
LTN reports on an older V100 GPU.

\begin{table}[h]
\centering
\caption{Test sum accuracy (\%), mean\,$\pm$\,std, on MNISTAdd for $N$ digits.  \nesycattorch averages $15$ seeds; LTN averages its $10$
\emph{best} of $15$. ${}^1$ from \citep{badreddineLogLTN2023}, ${}^2$ from \citep{kriekenANeSI2023}, ``T/O'': timeout; ``-'': not reported}
\label{tab:addition}
\small
\begin{tabular}{@{}lrccccc@{}}
                                                          &                         & \textbf{$N{=}1$}        & \multicolumn{2}{c}{\textbf{$N{=}2$}} & \textbf{$N{=}4$}                 \\
  \cmidrule(lr){3-3}\cmidrule(lr){4-5}\cmidrule(lr){6-6}
  \textbf{Method}                                         & \textbf{Trainings:}     & $3$k                    & $1.5$k                               & $15$k                   & $7.5$k \\
  \midrule
  \multicolumn{2}{l}{ \textbf{\nesycattorch (LTN-style)}} & $\mathbf{94.6 \pm 0.6}$ & $\mathbf{89.7 \pm 0.7}$ & $95.7 \pm 0.5$                       & $92.0 \pm 0.6$                   \\
  \multicolumn{2}{l}{\textbf{\nesycattorch (DPL-style)}}  & $94.2 \pm 0.7$    & $89.2 \pm 0.7$    & $95.8 \pm 0.6$                 & $91.8 \pm 0.8$                   \\
  \multicolumn{2}{l}{LTN}                                 & $93.5 \pm 0.3$ ${}^1$   & $88.4 \pm 1.0$ ${}^1$   & $95.4 \pm 0.3$          ${}^1$       & T/O                     ${}^2$         \\
  \multicolumn{2}{l}{logLTN}                              & -                       & $88.3 \pm 0.8$ ${}^1$   & $95.6 \pm 0.5$          ${}^1$       & -                                      \\
  \multicolumn{2}{l}{DeepProbLog}                         & $92.2 \pm 1.6$ ${}^1$   & $87.2 \pm 1.9$ ${}^1$   & $95.2 \pm 1.7$          ${}^2$       & T/O                     ${}^2$         \\
  \multicolumn{2}{l}{DeepStochLog}                        & -                       & -                       & $\mathbf{96.4 \pm 0.1}$ ${}^2$       & $\mathbf{92.7 \pm 0.6}$ ${}^2$         \\
  \multicolumn{2}{l}{A-NeSI}                              & -                       & -                       & $96.0 \pm 0.4$          ${}^2$       & $92.6 \pm 0.8$          ${}^2$         \\
  \midrule
  \emph{Reference} ($0.99^{2N}$)                          &                        & $98.01$ ${}^2$                      & $96.06$ ${}^2$                        & $96.06$ ${}^2$ & $92.27$ ${}^2$
\end{tabular}
\end{table}

\paragraph{Longer numbers.} The $N{=}4$ column uses two four-digit numbers (sums up to $19{,}998$). How
high any method can score here is set by the digit classifier: if it is $99\%$ accurate, getting all
$2N$ digits right caps the sum accuracy at $0.99^{2N}$ (the \emph{Reference} row). \nesycattorch, DeepProbLog and (in the non-recursive and mass-conserving case) also DeepStochLog
optimise the same marginal likelihood, so they end up near that cap. \nesycattorch
still runs at $N{=}4$ ($92.0\%$, just below the cap) where DeepProbLog and LTN time out, with
DeepStochLog and A-NeSI a little ahead.

\section{Related Work}
\label{sec:related}

\textbf{ULLER} \citep{vankriekenULLER2024} leaves the
semantics split into duplicated inductive definitions. The categorical framework of
\citet{schellhornNeSyCatCategorical2026} instead derives one definition parametric in a monad. But they do not implement neural learning, which we do here with \textbf{\nesycattorch}. \textbf{LTN}
\citep{DBLP:journals/ai/BadreddineGSS22} uses \emph{fuzzy} semantics and thereby avoid probabilistic semantics and hence do not derive any distributions.
Further, \textbf{logLTN} \citep{badreddineLogLTN2023} moves LTN's operators to log space
by composing $\log$ with softmax, whereas here log space is the native carrier and the
softmax is confined to the decode bridge.

\textbf{DeepProbLog} \citep{DBLP:journals/ai/ManhaeveDKDR21} trains on the same exact likelihood we do,
but compiles the program to a Sentential Decision Diagram and scores it by weighted model counting;
\citet{kriekenANeSI2023} show this counting is \#P-hard, so it times out at $N{=}4$
(Table~\ref{tab:addition}). \textbf{DeepStochLog} \citep{wintersDeepStochLog2022}  is the most accurate baseline, marginally ahead of both A-NeSI and \nesycattorch. It
 avoids DeepProbLog's blow-up by writing the sum as a stochastic grammar (a random walk rather than a random graph) which is cheaper but can lose mass on failing derivations. However, DeepStochLog changes the aggregation structure to grammar-derivation probability, while our approach keeps the distribution marginal known from DeepProbLog, yet achieves scalability via laziness and tensor structure. DeepStochLog's grammar-based approach corresponds to formulas dynamically adapted to the input (e.g.\ input length); we can achieve this by using \nesycattorch formulas inside Python or Haskell programs (that e.g.\ adapt the number of \textbf{do} binds to the number of input digits).

 \textbf{A-NeSI} \citep{kriekenANeSI2023}
scales instead by training a network to approximate the counting, which gives up exactness and requires additional factorization preparations, which are example specific and therefore not generalizable. \nesycattorch
needs none of this: the same marginalisation is just the monadic bind and by changing the monad the one framework also gives the classical and fuzzy semantics, more aligned with standard first-order logic.
 The
differentiable reading connects to the logic of differentiable logics
\citep{slusarzDifferentiableLogics2023} and to categorical deep learning
\citep{gavranovicCategoricalDeepLearning2024,fongBackpropFunctor2019}, with broader motivation in the
survey of \citet{smetDeepSeaProbLog2023}. 

\section{Conclusion}
\label{sec:conclusion}

NeSyCat is a general unifying neurosymbolic framework for reasoning and learning in first-order logic that is parameterized over a computational monad. Monads in NeSyCat have covered different theoretical frameworks, such as discrete and continuous probabilities and nondeterminism. With NeSyCat Torch, we provide the first neural implementation of this general framework, i.e.\ using neural networks as realisations of computational function and predicate symbols. We concentrate on finitely supported distributions here. We provide implementation at Haskell: \url{https://anonymous.4open.science/r/nesycattorch-hs/}, Python JAX: \url{https://anonymous.4open.science/r/nesycattorch-jax/}, and Python PyTorch: \url{https://anonymous.4open.science/r/nesycattorch-py/}.

It turns out that for these implementations, we need the finitely supported distribution monad plus two other monads: for (log-scale) tensors and for training batches. NeSyCat Torch thus scales the generality of NeSyCat to the neural implementation level, while simultaneously obtaining competitive results, in terms of accuracy and training time, for the classical MNIST digit addition example. Our approach also scales to multi-digit addition. This is achieved by using lazy monads that defer evaluation of marginalized probabilities to cases where really needed. Future work will study how well this generalises to more complex examples. 
The generalization of our implementation to other monads like continuous probability and to infinite domains is left to future work as well. Note that an efficient neural representation of continuous probability is non-trivial. 


\bibliography{../../daniel_zotero, ../../additional_citations}

\newpage
\appendix
\section{Categorical Background}\label{app:cat}

\paragraph{Monads.} Categorically, a monad on a category $\mathcal C$ is a functor $T \colon \mathcal C \to
\mathcal C$ with natural transformations $\eta \colon \mathrm{id}_{\mathcal C} \Rightarrow
T$ (\emph{unit}) and $\mu \colon TT \Rightarrow T$ (\emph{multiplication}) satisfying
$\mu \comp \eta T = \mathrm{id} = \mu \comp T\eta$ and $\mu \comp T\mu = \mu \comp \mu T$. The programming definition above corresponds
one-to-one to the equivalent presentation as a \emph{Kleisli triple} $(T, \eta, (\cdot)^{\mathcal M})$,
where $f^{\mathcal M} \colon TA \to TB$ for $f \colon A \to TB$ satisfies $\eta_A^{\mathcal M} =
\mathrm{id}_{TA}$, $f^{\mathcal M} \comp \eta_A^{\mathcal M} = f$, and $g^{\mathcal M} \comp f^{\mathcal M} = (g^{\mathcal M} \comp f)^{\mathcal M}$: \minth{return} is $\eta$, \minth{(>>=)} is the
Kleisli lift $(\cdot)^{\mathcal M}$ (applied flipped), and the \doNotation of Section~\ref{sec:monads}.
is its syntactic sugar.

\paragraph{States.} We work over a \emph{concrete} Cartesian category $\mathcal{C}$: objects are sets equipped
with structure (for example measurable spaces, tensor spaces or also plain sets), morphisms are
structure-preserving maps, finite products exist, and the terminal object $1$ is the
one-element set. Effectful maps are always written explicitly as maps
$f \colon S \to \mathcal M\,T$ of the chosen strong monad $\mathcal M$.
Because $\mathcal{C}$ is concrete and Cartesian, a \emph{state} on $S$, formally a Kleisli
point $1 \to \mathcal M\,S$, is the same thing as an \emph{element} of $\mathcal M\,S$; we
use this identification throughout and simply write $D \in \mathcal M\,S$.

\begin{proposition}[Pointwise evaluation]
  \label{prop:pointwise}
  For each $i \in \underline B$, evaluation $\mathrm{ev}_i \colon \mathcal B\mathcal M\,X \to
  \mathcal M\,X$, $m \mapsto m(i)$, is a monad morphism, and therefore commutes with the
  interpretation of \doNotation programs: for a batch $s \colon \underline B \to S$,
  \[
    \sem{\varphi}^{\mathcal B\mathcal M}(s)(i) \;=\; \sem{\varphi}^{\mathcal M}(s_i)
    \qquad (i \in \underline B).
  \]
  \end{proposition}
  
  \noindent
  One batched run thus yields all per-sample truth values; over the full
  sample these are the $N$ numbers $\widehat L$ averages, a mini-batch gives an unbiased
  estimate of $\widehat L$.
\end{document}